\documentclass{article}
\usepackage[preprint]{colm2026_conference}

\usepackage{microtype}
\usepackage{hyperref}
\usepackage{url}
\usepackage{booktabs}
\usepackage{lineno}
\usepackage{amsmath,amssymb,amsfonts}
\usepackage{graphicx}
\usepackage{multirow}
\usepackage{xcolor}
\usepackage{textgreek}
\usepackage{colortbl}
\usepackage{scrextend}

\usepackage{algorithm}
\usepackage{algorithmic}

\usepackage{amsthm}
\usepackage{wrapfig}
\usepackage{tikz}
\usepackage{pgfplots}
\usepackage{subcaption}
\usepackage{xcolor}
\pgfplotsset{compat=1.18}

\definecolor{oscarblue}{RGB}{76,114,176}
\definecolor{oscarorange}{RGB}{221,132,82}
\definecolor{oscargreen}{RGB}{85,168,104}
\definecolor{oscarred}{RGB}{196,78,82}
\definecolor{oscarpurple}{RGB}{129,114,179}
\definecolor{oscarbrown}{RGB}{147,120,96}
\definecolor{oscargray}{RGB}{120,120,120}
\definecolor{oscarlight}{RGB}{230,230,230}
\newtheorem{definition}{Definition}

\definecolor{darkblue}{rgb}{0,0,0.5}
\definecolor{lightgreen}{rgb}{0.8,1,0.8}
\definecolor{lightblue}{rgb}{0.8,0.9,1}
\hypersetup{
    colorlinks=true,
    citecolor=darkblue,
    linkcolor=darkblue,
    urlcolor=darkblue
}

\newcommand{\method}{\textsc{Oscar}}
\newcommand{\dlm}{DLM}
\newcommand{\dlms}{DLMs}

\newcommand{\ccentdist}{\mathcal{H}_{\times}}
\definecolor{oscarblue}{RGB}{0,112,192}
\definecolor{lightgreen}{RGB}{198,239,206}
\definecolor{lightblue}{RGB}{214,228,240}
\definecolor{lightyellow}{RGB}{255,242,204}
\definecolor{lightred}{RGB}{255,220,220}

\title{\method: Orchestrated Self-verification and Cross-path \\ Refinement}

\author{
Yash Shah\thanks{Equal contribution.} \\
Arizona State University \\
\texttt{yshah124@asu.edu}
\And
Abhijit Chakraborty\footnotemark[1] \\
Arizona State University \\
\texttt{achakr40@asu.edu}
\And
Naresh Kumar Devulapally \\
University at Buffalo, SUNY \\
\texttt{devulapa@buffalo.edu}
\And
Vishnu Lokhande \\
University at Buffalo, SUNY \\
\texttt{vishnulo@buffalo.edu}
\And
Vivek Gupta \\
Arizona State University \\
\texttt{vgupt140@asu.edu}
}

\begin{document}

\ifdefined\ifcolmsubmission
    \ifcolmsubmission
        \linenumbers
    \fi
\fi

\maketitle

\begin{abstract}
Diffusion language models (\dlms) expose their denoising trajectories, offering a natural handle for inference-time control; accordingly, an ideal hallucination mitigation framework should intervene during generation using this model-native signal rather than relying on an externally trained hallucination classifier.
Toward this, we formulate \emph{commitment uncertainty localization}: given a
denoising trajectory, identify token positions whose cross-chain entropy exceeds
an unsupervised threshold before factually unreliable commitments propagate into
self-consistent but incorrect outputs.
We introduce a suite of trajectory-level assessments, including a cross-chain
divergence-at-hallucination (CDH) metric, for principled comparison of
localization methods.
We also introduce \method, a training-free inference-time framework
operationalizing this formulation.
\method~ runs $N$ parallel denoising chains with randomized reveal orders,
computes cross-chain Shannon entropy to detect high-uncertainty positions,
and then performs targeted remasking conditioned on retrieved evidence.
Ablations confirm that localization and correction contribute complementary
gains, robust across $N \in \{4, 8, 16\}$.
On TriviaQA, HotpotQA, RAGTruth, and CommonsenseQA using LLaDA-8B and
Dream-7B, \method~ enhances generation quality by significantly reducing hallucinated content and improving factual accuracy through uncertainty-guided remasking, which also facilitates more effective integration of retrieved evidence. Its native entropy-based uncertainty signal surpasses that of specialized trained detectors, highlighting an inherent capacity of diffusion language models to identify factual uncertainty that is not present in the sequential token commitment structure of autoregressive models. 

\end{abstract}


\section{Introduction}
The Transformer architecture \citep{vaswaniAttentionAllYou2017} revolutionized sequence modeling by replacing recurrence with self-attention, leading to autoregressive (AR) decoding for token generation. This enables hallucination detection tools like perplexity thresholding and retrieval-augmented verification \citep{farquharDetectingHallucinationsLarge2024,malininUncertaintyEstimationAutoregressive2021,lewisRetrievalAugmentedGenerationKnowledgeIntensive2020,tonmoyComprehensiveSurveyHallucination2024}. Diffusion language models (\dlms) generate text through iterative demasking from masked sequences \citep{nieLargeLanguageDiffusion2025,yeDream7BDiffusion2025}. Unlike autoregressive LLMs, \dlms{} maintain a trajectory \citep{liSurveyDiffusionLanguage2025} of intermediate states encoding uncertainty, impacting question answering where early errors cause hallucination crystallization \citep{luUnderstandingTextHallucination2024}. These research threads hallucination detection in language models and denoising dynamics of diffusion models developed largely in parallel. Detection literature focuses on classifiers recognizing hallucinations post-generation, while \dlm literature characterizes trajectory-level phenomena without exploiting them for correction. These threads converge: \dlm denoising's order-dependence of token commitment directly yields an uncertainty signal that detection literature trains classifiers to approximate \citep{bloem-reddyProbabilisticSymmetriesInvariant2020}. \method~ bridges this gap by transforming a diagnostic signal into a correction mechanism.\newline\indent Prior work on hallucination detection includes \emph{output-based} methods\citep{kuhnSemanticUncertaintyLinguistic2023,linGeneratingConfidenceUncertainty2023,renOutofDistributionDetectionSelective2023,malininUncertaintyEstimationAutoregressive2021} analyzing final output, \emph{latent-based} methods\citep{burnsDiscoveringLatentKnowledge2024,parkSteerLLMLatents2025,azariaInternalStateLLM2023,chenINSIDELLMSInternal2024} probing internal representations, and \emph{trajectory-based} methods TraceDet\citep{changTraceDetHallucinationDetection2025}, DynHD\citep{qianDynHDHallucinationDetection2026} exploiting multi-step dynamics. Trajectory-based methods achieve strongest results but require trained classifiers that may not transfer across domains. These methods only flag hallucinations without correction. This is due to AR-based tools' incompatibility with bidirectional denoising, lack of ground-truth signals, and underutilized trajectory data.\citep{niuRAGTruthHallucinationCorpus2024,minFActScoringFineGrainedAtomic2023,tonmoyComprehensiveSurveyHallucination2024}.\newline\indent We argue that rather than training classifiers to recognize hallucinations after generation, we should exploit the \emph{native uncertainty signal} from \dlms{}. Our key observation is: \begin{quote}\emph{If multiple denoising chains, each following a different randomized reveal order, disagree on a token position, that position is factually uncertain; regardless of how confident any single chain appears.}\end{quote}
\begin{figure}[htbp]
    \centering  \includegraphics[width=\linewidth, height=5cm]{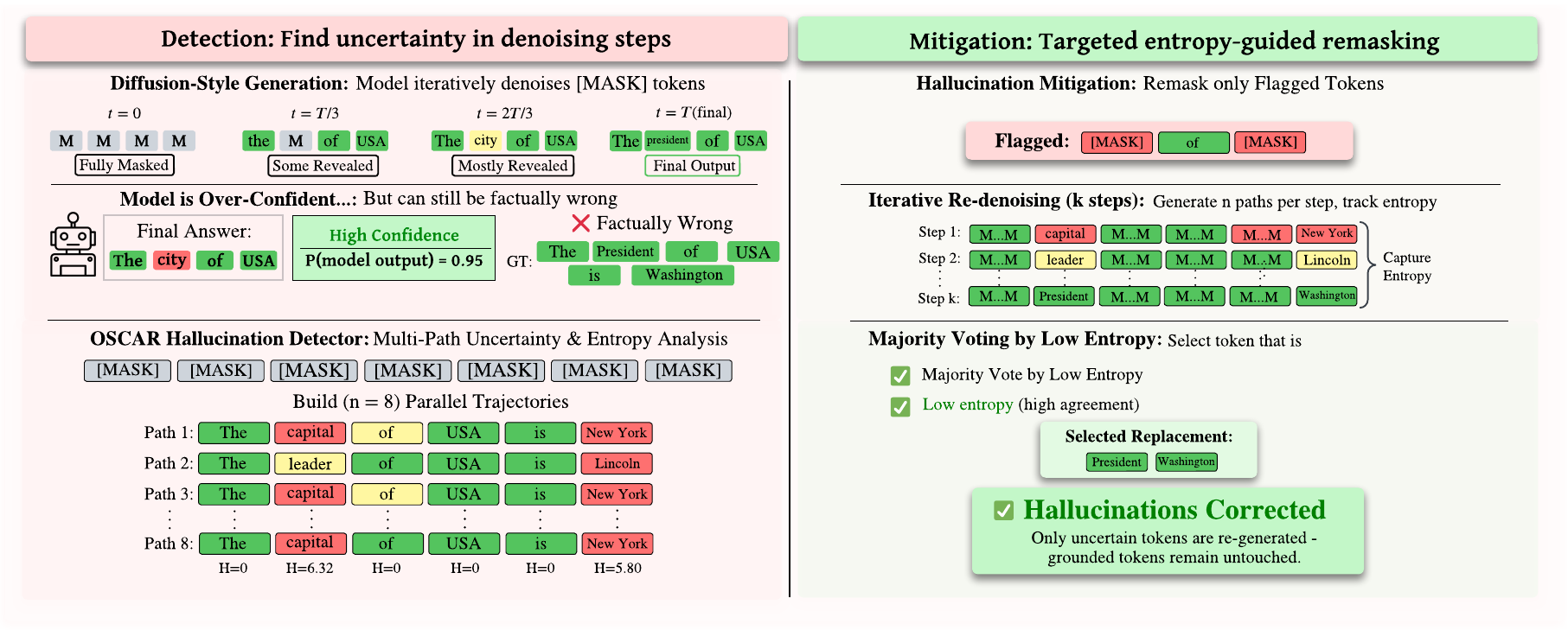}
    \caption{Schematic overview of the \method~ pipeline, illustrating the key stages and components involved in the process.}
    \label{fig:overview}
\end{figure}
This leads to \textbf{commitment uncertainty localization} (\S\ref{sec:formulation}): given parallel denoising trajectories, identify token positions where cross-chain Shannon entropy exceeds an unsupervised threshold, \emph{before} premature commitments lead to incorrect outputs. The \textbf{Adaptive Remasking and Re-denoising Phase} (\S\ref{sec:formulation}) then selectively remasks uncertain token spans and re-denoises them, enabling targeted correction based on uncertainty signal. We implement these in \method{} (\textbf{O}rchestrated \textbf{S}elective \textbf{C}orrection via \textbf{A}daptive \textbf{R}emasking), a training-free inference-time framework. \method{} runs $N$ parallel denoising chains with randomized reveal orders, computes cross-chain entropy to localize uncertainties, and performs targeted remasking with retrieved evidence. The pipeline requires only modest runtime increase while delivering substantial quality gains. \method~ is positioned as the first training-free zero-shot \dlm~ native hallucination detection and mitigation method.
\par\indent Our work fully characterizes commitment uncertainty in diffusion language models, its origin, measurement, actions, and formation, organized into four contributions: (a) We identify and demonstrate a native uncertainty signal inherent in diffusion language models (\dlms{}), derived from cross-chain Shannon entropy computed across multiple parallel denoising trajectories with randomized reveal orders. This signal enables competitive hallucination detection without external supervision, highlighting an uncertainty measure structurally absent in autoregressive models. (b) We formalize the task of commitment uncertainty localization by introducing the cross-chain divergence-at-hallucination (CDH) metric, which effectively identifies factually unreliable token positions within denoising trajectories, substantially outperforming baseline methods. (c) We propose \method, the first inference-time hallucination mitigation framework for \dlms, which leverages targeted re-masking and re-denoising of uncertain token spans to correct hallucinations. \method~ achieves significant improvements in generation quality and factual accuracy across multiple benchmarks while maintaining efficient computational overhead. (d) We provide a mechanistic analysis of hallucination crystallization timing during the denoising process, revealing task-dependent patterns that support early-exit detection strategies and enable adaptive, task-specific intervention.

\section{Related Work}
\label{sec:related}
\par\textbf{Hallucination detection in autoregressive (AR) language models} has developed through three main approaches. Uncertainty-based methods estimate confidence by examining output token distributions \citep{malininUncertaintyEstimationAutoregressive2021, renOutofDistributionDetectionSelective2023}. Semantic entropy methods group equivalent outputs before calculating entropy to capture meaning-level uncertainty \citep{kuhnSemanticUncertaintyLinguistic2023, farquharDetectingHallucinationsLarge2024}. Probe-based methods train lightweight classifiers on internal model activations to identify hallucinations \citep{azariaInternalStateLLM2023, burnsDiscoveringLatentKnowledge2024}. SelfCheckGPT evaluates agreement across multiple generations but requires multiple generations and lacks correction mechanisms \citep{manakulSelfCheckGPTZeroResourceBlackBox2023}. Evaluation has evolved, with ROUGE-based metrics shown inferior to LLM-as-Judge protocols, leading to combined evaluation approaches \citep{janiakIllusionProgressReevaluating2025}. However, current methods do not address hallucination detection in diffusion language models (\dlms), which use parallel bidirectional denoising architecture, necessitating new \dlm-specific approaches.
\par\textbf{Hallucination detection in diffusion language models} (\dlms) has recently gained attention as a promising area of research. Methods such as TraceDet model the denoising process as an action trace and identify the most informative sub-traces using the Information Bottleneck principle, resulting in a 15.2\% average AUROC improvement over output-based baselines. DynHD tackles the issue of information density imbalance across token positions by constructing semantic-aware evidence and introducing a reference evidence generator that learns the expected uncertainty evolution, detecting hallucinations by measuring deviations from this reference. EigenScore \citep{shoushtariEigenScoreOODDetection2025} leverages the eigenvalue spectrum of the posterior covariance induced by the diffusion model as an out-of-distribution (OOD) detection signal. While these approaches effectively use the denoising trajectory as diagnostic input for classification, they do not modify or control the denoising process itself. This limitation hinders adaptability, as the denoising trajectory remains a passive signal rather than an active means for correction. 
\par Existing approaches can be viewed as restricted instantiations of a general principle: \textit{disagreement among diverse generations reveals factual uncertainty}.  Independent resampling (SelfCheckGPT~\citep{manakulSelfCheckGPTZeroResourceBlackBox2023}) measures total output variance but conflates sampling noise with factual uncertainty because each resample traverses the full generation process independently.  Trajectory classifiers (TraceDet, DynHD) learn to recognize hallucination-correlated patterns in the denoising trace but require labeled data and provide no correction affordance.  Cross-chain entropy under reveal-order diversification isolates the~\textit{commitment-order} component of disagreement directly, the precise structural source of \dlm~ hallucination, without training.  \method~ operationalizes this signal end-to-end: it transforms the denoising trajectory from a passive diagnostic input into an active control mechanism, enabling targeted correction at the positions where commitment uncertainty is highest.

\section{\method: Commitment Uncertainty Localization and Correction}
\label{sec:formulation}

We present \method~ in four parts that mirror our contributions.
We first identify the structural failure mode underlying hallucination in \dlms~and show that it produces a measurable, model-native uncertainty signal
(\S\ref{sec:signal}; contribution~a).
We then formalize the localization problem and introduce the CDH evaluation
metric (\S\ref{sec:localization}; contribution~b).
We describe the full detection-and-correction pipeline
(\S\ref{sec:pipeline}; contribution~c), and finally connect the signal's
temporal structure to hallucination crystallization dynamics
(\S\ref{sec:crystallization_setup}; contribution~d).

\subsection{A Native Uncertainty Signal in Diffusion Language Models}
\label{sec:signal}

A \dlm~ generates response $\mathbf{y} = (y_1, \ldots, y_L)$ to query $q$
through $T$ denoising steps.
At $t{=}0$ all positions are masked; at each step the model predicts
$p_\theta(y_i \mid \mathbf{y}^{(t-1)}, q)$ and reveals a subset according to
a \textit{reveal order} $\pi$.
The resulting states form a \textit{denoising trajectory}
$\tau^{\pi} = (\mathbf{y}^{(0)}, \ldots, \mathbf{y}^{(T)})$.
\label{sec:trajectories}
Standard inference uses a confidence-based order;
LLaDA-8B~\citep{nieLargeLanguageDiffusion2025} uses $T{=}128$ steps,
Dream-7B~\citep{yeDream7BDiffusion2025} uses $T{=}64$.

Once revealed, a token is \textit{committed}: all subsequent steps condition
on it irreversibly.
A factually incorrect early commitment acts as a \textit{contextual attractor},
biasing neighbors toward internally consistent but wrong predictions via
bidirectional attention~\citep{changTraceDetHallucinationDetection2025,
qianDynHDHallucinationDetection2026}.
We call this \textit{premature commitment hallucination}.
Unlike autoregressive hallucination, confined to the generation frontier, a
single early error propagates through all remaining denoising steps.

The key insight is that such hallucinations are \textit{order-dependent}: a
different reveal order $\pi' \neq \pi$ may commit a different token first,
potentially yielding a correct output.
Running $N$ parallel chains with independently randomized reveal orders exposes
this latent uncertainty without external supervision.

\begin{definition}[Cross-Chain Entropy]
\label{def:cross_chain}
\label{sec:cross_entropy}
Let $y_i^{(T,n)}$ denote the final token at position $i$ from chain $n$.
The empirical distribution is
$\hat{p}_i(v) = \frac{1}{N}\sum_{n=1}^{N}
\mathbb{1}[y_i^{(T,n)} = v]$.
The \textbf{cross-chain entropy} at position $i$ is:
\begin{equation}
H_{\times,i} \;=\; -\!\sum_{v \in \mathcal{V}}
\hat{p}_i(v)\,\log\,\hat{p}_i(v)
\label{eq:cross_chain}
\end{equation}
\end{definition}

\noindent
Positions with $H_{\times,i}{=}0$ are \textit{commitment-stable}: every chain
agrees regardless of reveal order.
Positions with high $H_{\times,i}$ are \textit{commitment-uncertain}: the
output depends on the arbitrary ordering rather than on factual knowledge.
Position $i$ is commitment-stable if all reveal orders yield the same token: $H_{\times,i} = 0$ characterizes this. Cross-chain entropy is the complete measure of commitment disagreement: zero when order-invariant, positive when reveal schedule affects output. This signal is \textit{model-native}, from the \dlm's denoising dynamics without a trained classifier or labeled data, and \textit{structurally specific to \dlms}; autoregressive generation needs full resamples, while \dlms\ use parallel chains to isolate effects. With token vocabulary $|\mathcal{V}|$ over 32,000, the mean distinct tokens per position across $N{=}8$ chains is $2.7$ (median~$1$, max~$8$), ensuring reliable entropy estimation. Cross-chain entropy, as a standalone detection signal, outperforms trained trajectory classifiers, capturing factual uncertainty more directly (\S\ref{sec:results_detection}).

\subsection{Localization Formulation and CDH Metric}
\label{sec:localization}

We formalize the task of identifying factually unreliable positions from
parallel denoising trajectories.

\begin{definition}[Commitment Uncertainty Localization]
\label{def:localization}
Given query $q$ and $N$ trajectories $\{\tau^{\pi_n}\}_{n=1}^{N}$, identify:
\begin{equation}
\mathcal{U} \;=\;
\bigl\{\,i \;:\; H_{\times,i} > Q_{1-\alpha}(H_{\times,1:L})\,\bigr\}
\label{eq:localization}
\end{equation}
where $Q_{1-\alpha}$ is the $(1{-}\alpha)$-quantile of the entropy
distribution (default \emph{localization aggressiveness threshold}~$(\alpha){=}0.2$, flagging the top $20\%$ most uncertain
positions).
\end{definition}

\noindent
This formulation is (1)~\textbf{training-free}: $Q_{1-\alpha}$ is computed
from the current sample, requiring no labeled data;
(2)~\textbf{model-native}: the signal originates entirely from the \dlm's
denoising process; and
(3)~\textbf{pre-correction}: uncertainty is measured from $N$ parallel chains
before any modification, avoiding post-hoc circularity.

\label{sec:cdh}
To evaluate localization quality at the token level, we introduce the
\textit{cross-chain divergence-at-hallucination rate}:
\begin{equation}
\mathrm{CDH}(k) \;=\;
\frac{\bigl|\,\{i \in \mathcal{U}_k : i \text{ is hallucinated}\}\,\bigr|}
     {\bigl|\,\{i : i \text{ is hallucinated}\}\,\bigr|}
\label{eq:cdh}
\end{equation}
where $\mathcal{U}_k$ contains positions with the top-$k\%$ highest
$H_{\times,i}$.
CDH$(k)$ measures the fraction of truly hallucinated positions captured by the
top-$k\%$ most uncertain positions (random baseline: $k/100$).
Because \method's correction acts on localized spans, CDH directly measures the
signal driving mitigation performance.
We show in \S\ref{sec:cdh_analysis} that cross-chain entropy concentrates
hallucinated positions in the high-uncertainty tail far more effectively than
both random selection and the strongest trained alternative, confirming that
unsupervised localization from the model's own trajectories is sufficient for
targeted correction.

\subsection{The \method~Pipeline}
\label{sec:pipeline}
\label{sec:method}

\begin{figure}[t]
    \centering
    \includegraphics[width=1\linewidth]{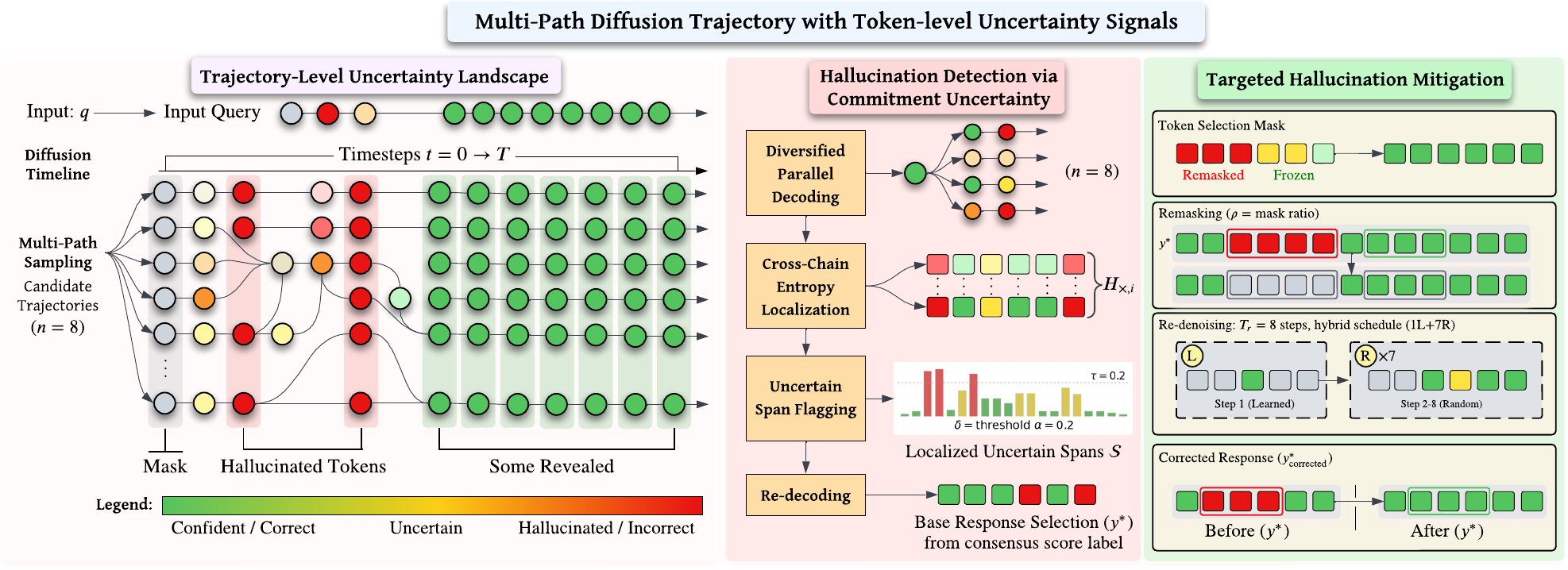}
    \caption{\method~ overview.
    Phase~1: $N$ parallel denoising chains with randomized reveal orders
    generate diverse trajectories.
    Cross-chain entropy $H_{\times}$ identifies high-uncertainty positions
    (red).
    Phase~2: targeted remasking, conditioned on retrieved evidence, corrects
    hallucinated spans at $1.3{\times}$ overhead.}
    \label{fig:overview}
\end{figure}
Before detailing the pipeline, note that \method~ implements a program for commitment uncertainty localization in \dlms:\textbf{(1)} diversify reveal orders to expose latent factual uncertainty;\textbf{(2)} compute a position-level disagreement statistic from trajectories;\textbf{(3)} threshold to localize uncertain spans;\textbf{(4)} correct via targeted re-denoising conditioned on context or evidence.\method~ uses cross-chain Shannon entropy as the disagreement statistic, but alternatives like token-level KL divergence or embedding-space distance are possible. The correction stage could use stronger evidence sources or iterative refinement. We present choices yielding the strongest empirical results; the general program is the conceptual contribution.\method~ operationalizes commitment uncertainty localization in three stages(Figure~\ref{fig:overview}).
\paragraph{Stage 1: Diversified parallel decoding.}
\label{sec:diversify}
\method~ samples $N$ reveal orders uniformly at random and runs $N$ chains in
parallel, sharing model weights~$\theta$.
Standard confidence-based decoding always commits the most confident predictions
first, suppressing observable uncertainty via contextual
attractors~(\S\ref{sec:signal}).
Randomized orders break this coupling, ensuring no position receives systematic
priority.
All chains are batched into a single forward pass per step, yielding approximately one and a  fifth times additional processing time.

\paragraph{Stage 2: Localization and span aggregation.}
\label{sec:localize}
\method~ computes $H_{\times,i}$ (Eq.~\ref{eq:cross_chain}) at each position and
applies the percentile threshold (Eq.~\ref{eq:localization}) to obtain the
uncertain set~$\mathcal{U}$.
Consecutive positions are grouped into spans, each extended by $w{=}2$ tokens
on either side; spans shorter than $\ell_{\min}{=}3$ tokens are discarded
(ablated in Appendix~\ref{app:span}).
The base response $\mathbf{y}^*$ is the chain whose tokens best match the
cross-chain consensus:
\begin{equation}
n^* \;=\; \arg\max_{n} \;\sum_{i=1}^{L}
\hat{p}_i\!\bigl(y_i^{(T,n)}\bigr)
\label{eq:base_selection}
\end{equation}

\paragraph{Stage 3: Targeted remasking and correction.}
\label{sec:correct}
For each span $s = (a, \ldots, b) \in \mathcal{S}$, \method remasks positions
$a$ through $b$ in $\mathbf{y}^*$ and runs $T_r$ denoising steps with all
surrounding tokens frozen:
\begin{equation}
\hat{\mathbf{y}}_s \;=\; \arg\max_{\mathbf{y}_{a:b}} \;
p_\theta\!\Bigl(\mathbf{y}_{a:b}
\;\Big|\;
\mathbf{y}^*_{\setminus s},\;
[\mathbf{e}_s;\,q]\Bigr)
\quad \text{via } T_r \text{ denoising steps}
\label{eq:correction}
\end{equation}
where $\mathbf{e}_s$ is optional retrieved evidence.
This exploits the DLM's native ability to fill masked regions given surrounding
context, an affordance unavailable in autoregressive models.

The first refinement step uses the learned confidence-based order, anchoring
the correction with the most confident prediction to prevent a new attractor.
The remaining $T_r{-}1$ steps use random orders, allowing the span to settle
without reproducing the original error 
When a retrieval corpus is available, the uncertain span text serves as the
query (rather than the full input), focusing evidence on the specific claim
under correction.
We use Contriever~\citep{izacardUnsupervisedDenseInformation2022} over a 2021
Wikipedia snapshot (top-1 passage, 256 tokens).

\paragraph{Overhead and hyperparameters.}
The correction stage remasks ${\sim}18.7\%$ of tokens on average.
With $T_r{=}8$ steps and batched inference, correction adds ${\sim}0.1{\times}$;
combined with diversification (${\sim}1.2{\times}$), total overhead is
${\sim}1.3{\times}$.
Defaults: $N{=}8$ chains, $\alpha{=}0.2$, $T_r{=}8$, $w{=}2$,
$\ell_{\min}{=}3$, 1-Learned~$+$~$(T_r{-}1)$-Random schedule.
Each is validated by ablations in later sections.

\subsection{Crystallization: When Do Hallucinations Form?}
\label{sec:crystallization_setup}

The cross-chain entropy signal is not only diagnostic of \textit{where}
hallucinations occur but also \textit{when} they form during denoising.
We define the \textit{entropy gap} at step $t$ as the difference in mean
cross-chain entropy between hallucinated and grounded positions:
\begin{equation}
\Delta H(t) \;=\; \mathbb{E}\!\bigl[H_{\times} \mid \text{hallucinated},\,t\bigr]
             \;-\; \mathbb{E}\!\bigl[H_{\times} \mid \text{grounded},\,t\bigr]
\label{eq:entropy_gap}
\end{equation}
We find $\Delta H(t)$'s temporal profile is task-dependent. In factual QA, the entropy gap peaks early and persists, showing factual uncertainty crystallizes quickly. For summarization, the gap grows mid-denoising, reflecting compositional errors. In factual QA, an early incorrect commitment dominates due to bidirectional attention. In summarization, hallucinations from compositional choices across spans require more denoising to detect cross-chain disagreement. Token-level ground truth uses RAGTruth's span annotations; for TriviaQA, we align the final gold answer against each intermediate state via exact-match at each step. This crystallization pattern supports \method~ as diverse reveal orders are informative when commitment-order effects are strongest, suggesting early-exit detection to reduce overhead (full analysis in~\S\ref{sec:analysis}).

\section{Experimental Setup}
\label{sec:setup}
 
\paragraph{Models and datasets.}
We evaluate on LLaDA-8B-Instruct~\citep{nieLargeLanguageDiffusion2025}
($T{=}128$ steps) and Dream-7B-Instruct~\citep{yeDream7BDiffusion2025}
($T{=}64$).
Benchmarks: TriviaQA~\citep{joshiTriviaQALargeScale2017} (500),
HotpotQA~\citep{yangHotpotQADatasetDiverse2018} (500),
CommonsenseQA~\citep{talmorCommonsenseQAQuestionAnswering2019} (500; negative
control---multiple-choice format yields $H_\times{\approx}0$), and
RAGTruth~\citep{niuRAGTruthHallucinationCorpus2024} (500; token-level span
annotations).
 
\paragraph{Baselines and evaluation.}
Nine baselines across three families:
output-based (Perplexity, LN-Entropy, Semantic Entropy%
\citep{kuhnSemanticUncertaintyLinguistic2023,farquharDetectingHallucinationsLarge2024},
Lexical Similarity\citep{manakulSelfCheckGPTZeroResourceBlackBox2023}),
latent-based (EigenScore\citep{shoushtariEigenScoreOODDetection2025},
CCS\citep{burnsDiscoveringLatentKnowledge2024},
TSV\citep{changTraceDetHallucinationDetection2025}),
and trajectory-based (TraceDet\citep{changTraceDetHallucinationDetection2025},
DynHD\citep{qianDynHDHallucinationDetection2026}),
plus token-level majority vote across $N\!=\!8$ chains.
We report AUROC under EM and LLM-as-Judge (GPT-4o\citep{openaiGPT4TechnicalReport2024}), F1 before/after correction,
CDH($k$), and span reduction on RAGTruth\citep{niuRAGTruthHallucinationCorpus2024};
all as mean $\pm$ std over three seeds.
Defaults: $N\!=\!8$, $\alpha\!=\!0.2$, $T_r\!=\!8$; batched on $4\times$H200 80GB;
retrieval via Contriever\citep{izacardUnsupervisedDenseInformation2022} over Wikipedia.

\section{Discussion}
\label{sec:results}
\subsection{Hallucination Detection: \method{} Surpasses Trained Detectors}
\label{sec:results_detection}

Table~\ref{tab:main} presents AUROC results. Under LLM-as-Judge evaluation, \method{} achieves \textbf{86.5\%} average AUROC on LLaDA-8B and \textbf{85.7\%} on Dream-7B---surpassing DynHD (the strongest trained detector) by \textbf{+2.3} and \textbf{+1.4} points respectively. Under exact-match evaluation, \method{} scores 76.4\% (LLaDA-8B), which is lower than DynHD's 84.2\% because EM penalizes \method{}'s semantically correct but lexically different corrections.
 \begin{wraptable}{r}{0.7\linewidth}
\centering
\footnotesize
\setlength{\tabcolsep}{3pt}
\begin{tabular}{@{}llccccccr@{}}
\toprule
& & \multicolumn{2}{c}{\textbf{TriviaQA}} & \multicolumn{2}{c}{\textbf{HotpotQA}} & \multicolumn{2}{c}{\textbf{CSQA}} & \\
\cmidrule(lr){3-4} \cmidrule(lr){5-6} \cmidrule(lr){7-8}
& \textbf{Method} & 128 & 64 & 128 & 64 & 128 & 64 & \textbf{Avg} \\
\midrule
\multicolumn{9}{@{}l}{\textit{LLaDA-8B-Instruct}} \\
\midrule
\multirow{4}{*}{\rotatebox[origin=c]{90}{\scriptsize Output}} & Perplexity & 50.4 & 47.6 & 49.3 & 51.2 & 65.6 & 65.0 & 54.9 \\
& LN-Entropy & 54.6 & 53.5 & 54.8 & 54.7 & 64.6 & 64.4 & 57.8 \\
& Semantic Entropy & 68.9 & 67.3 & 57.6 & 53.8 & 44.1 & 43.9 & 55.9 \\
& Lexical Similarity & 62.5 & 59.0 & 64.2 & 57.1 & 57.3 & 60.7 & 60.1 \\
\midrule
\multirow{3}{*}{\rotatebox[origin=c]{90}{\scriptsize Latent}} & EigenScore & 69.2 & 66.9 & 64.7 & 59.2 & 58.5 & 60.6 & 63.2 \\
& CCS & 57.1 & 54.2 & 57.6 & 55.8 & 50.5 & 58.5 & 55.6 \\
& TSV & 60.2 & 61.1 & 65.0 & 59.4 & 52.9 & 55.2 & 59.0 \\
\midrule
\multirow{4}{*}{\rotatebox[origin=c]{90}{\scriptsize Traj.}} & TraceDet$^\dagger$ & 73.9 & 74.1 & 66.1 & 63.7 & 77.2 & 77.1 & 72.0 \\
& DynHD$^\dagger$ & \underline{86.7} & \underline{86.1} & \underline{84.2} & \underline{85.3} & \underline{81.6} & \underline{81.3} & \underline{84.2} \\
& \cellcolor{lightgreen}\method{} (EM) & \cellcolor{lightgreen}80.3 & \cellcolor{lightgreen}79.8 & \cellcolor{lightgreen}71.5 & \cellcolor{lightgreen}68.2 & \cellcolor{lightgreen}79.4 & \cellcolor{lightgreen}78.9 & \cellcolor{lightgreen}76.4 \\
& \cellcolor{lightblue}\method{} (Judge) & \cellcolor{lightblue}\textbf{89.7} & \cellcolor{lightblue}\textbf{88.8} & \cellcolor{lightblue}\textbf{86.7} & \cellcolor{lightblue}\textbf{87.5} & \cellcolor{lightblue}\textbf{83.4} & \cellcolor{lightblue}\textbf{82.8} & \cellcolor{lightblue}\textbf{86.5} \\
\midrule\midrule
\multicolumn{9}{@{}l}{\textit{Dream-7B-Instruct}} \\
\midrule
\multirow{2}{*}{\rotatebox[origin=c]{90}{\scriptsize Out.}} & Semantic Entropy & 73.7 & 72.5 & 62.7 & 67.7 & 51.4 & 48.6 & 62.8 \\
& Lexical Similarity & 58.3 & 64.0 & 59.7 & 62.7 & 77.3 & 76.9 & 66.5 \\
\midrule
\multirow{3}{*}{\rotatebox[origin=c]{90}{\scriptsize Lat.}} & EigenScore & 66.0 & 69.1 & 62.5 & 67.0 & 76.9 & 77.5 & 69.8 \\
& CCS & 56.9 & 50.3 & 51.7 & 58.2 & 54.2 & 53.2 & 54.1 \\
& TSV & 75.6 & 74.7 & 58.7 & 63.0 & 62.3 & 56.8 & 65.2 \\
\midrule
\multirow{4}{*}{\rotatebox[origin=c]{90}{\scriptsize Traj.}} & TraceDet$^\dagger$ & 78.1 & 86.7 & 75.1 & 76.0 & 84.7 & 84.1 & 80.8 \\
& DynHD$^\dagger$ & \underline{87.3} & 84.4 & \underline{80.1} & \underline{85.6} & 83.5 & 84.6 & \underline{84.3} \\
& \cellcolor{lightgreen}\method{} (EM) & \cellcolor{lightgreen}82.6 & \cellcolor{lightgreen}83.1 & \cellcolor{lightgreen}77.4 & \cellcolor{lightgreen}79.8 & \cellcolor{lightgreen}84.2 & \cellcolor{lightgreen}83.7 & \cellcolor{lightgreen}81.8 \\
& \cellcolor{lightblue}\method{} (Judge) & \cellcolor{lightblue}\textbf{89.5} & \cellcolor{lightblue}\textbf{86.3} & \cellcolor{lightblue}\textbf{81.9} & \cellcolor{lightblue}\textbf{87.1} & \cellcolor{lightblue}\textbf{84.3} & \cellcolor{lightblue}\textbf{85.1} & \cellcolor{lightblue}\textbf{85.7} \\
\bottomrule
\multicolumn{9}{@{}l}{\scriptsize $^\dagger$ = requires trained classifier on hallucination labels. Also uses LLM-as-Judge eval}
\end{tabular}
\caption{\small\textbf{AUROC(\%) on two \dlms{} across three QA datasets.} Best \textbf{bolded}, second \underline{underlined}. \colorbox{lightgreen}{Green} = \method{} (EM eval). \colorbox{lightblue}{Blue} = \method{} (LLM-as-Judge eval). All trained baselines use EM labels only.}
\label{tab:main}
\vspace{-3em}
\end{wraptable}
The \textbf{+10.7 AUROC gain} reflects the gap between \method{}'s raw entropy signal-localization only, without correction, and TraceDet (82.7 vs. 72.0 on LLaDA-8B).\footnote{The entropy-only AUROC uses cross-chain entropy $\ccentdist$ as the detection score directly, without the correction stage. With correction and LLM-as-Judge relabeling, the full-pipeline AUROC rises further to 86.5\%.} This gain is achieved \emph{without any hallucination-specific training}, revealing that the cross-chain entropy signal captures factual uncertainty more effectively than trained trajectory classifiers.
\newline
\subsection{Generation Quality and Span Correction}
\label{sec:results_generation}
 

Figure~\ref{fig:generation} shows F1 scores before and after correction. OSCAR improves QA macro-average F1 by $+6.1$\,pp on LLaDA-8B and $+8.3$\,pp across both models, with the largest gain in TriviaQA ($+10.7$). In CommonsenseQA, $\Delta$F1$\,{=}\,$0.0 as chains agree on single-token answers ($H_\times{\approx}0$), so OSCAR doesn't intervene. The strong sample-level AUROC on CommonsenseQA shows high confidence outputs ($H_\times{\approx}0$) are factual; position-level 
\begin{figure}[htbp]  
\centering  
\begin{minipage}[t]{0.48\textwidth}    
\centering    
\includegraphics[width=\linewidth]{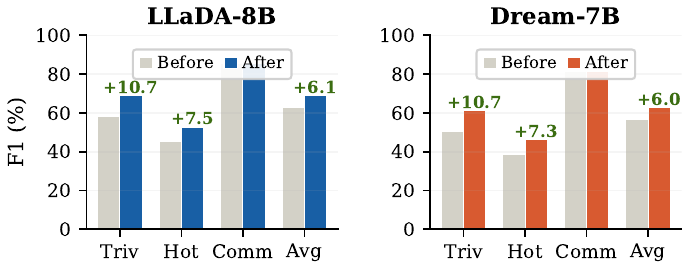}    \caption{\small F1 before (gray) and after (color) OSCAR correction.    Green annotations show $\Delta$F1.    CommonsenseQA shows no change ($H_\times{\approx}0$), validating    selectivity.    Mean $\pm$ std over 3 seeds; full metrics in Appendix~\ref{app:generation}.}    \label{fig:generation}  \end{minipage}  \hfill  \begin{minipage}[t]{0.48\textwidth}    \centering    \includegraphics[width=\linewidth]{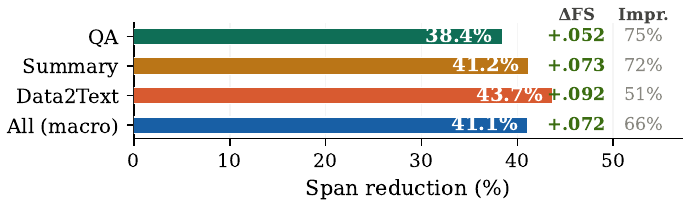}    \caption{\small Span-level correction on RAGTruth (LLaDA-8B).    Bars show hallucinated span reduction; $\Delta$FS and \% Improved    annotated at right.    Full per-subset metrics in Appendix~\ref{app:ragtruth}.}    
\label{fig:ragtruth}  
\end{minipage}\end{figure} 
$H_\times{\approx}0$ indicates no uncertain spans to correct, these signals are complementary, not contradictory. Results for Dream-7B show similar improvements. Figure~\ref{fig:ragtruth} shows OSCAR's correction extends beyond short-form QA: hallucinated spans decrease 38--44\% across RAGTruth subsets, with positive $\Delta$FS showing improved factual precision. Uncertainty-targeted retrieval adds $+3.4$ F1 over OSCAR (Appendix~\ref{app:retrieval}).

\subsection{Localization Quality: CDH Analysis}
\label{sec:cdh_analysis}
 
Appendix~\ref{app:cdh} reports CDH($k$) curves. At $k{=}20\%$ (i.e., examining only the top 20\% most uncertain positions), \method{}'s cross-chain entropy captures 67.3\% of all hallucinated positions on LLaDA-8B---more than $3\times$ the random baseline (20\%) and 1.4$\times$ the best trained detector (TraceDet, 47.8\%). This validates that cross-chain entropy is a highly effective localization signal without requiring any supervision.
\subsection{Pareto Analysis: Performance vs.\ Efficiency}
Figure~\ref{fig:pareto_data} presents the Pareto frontier. \method{} (with Judge evaluation) achieves the highest AUROC at $1.3\times$ wall-clock overhead, establishing a new Pareto-optimal point that encompasses both hallucination mitigation and detection.
\begin{wrapfigure}{r}{0.5\textwidth}
    \centering
    \includegraphics[width=0.48\textwidth]{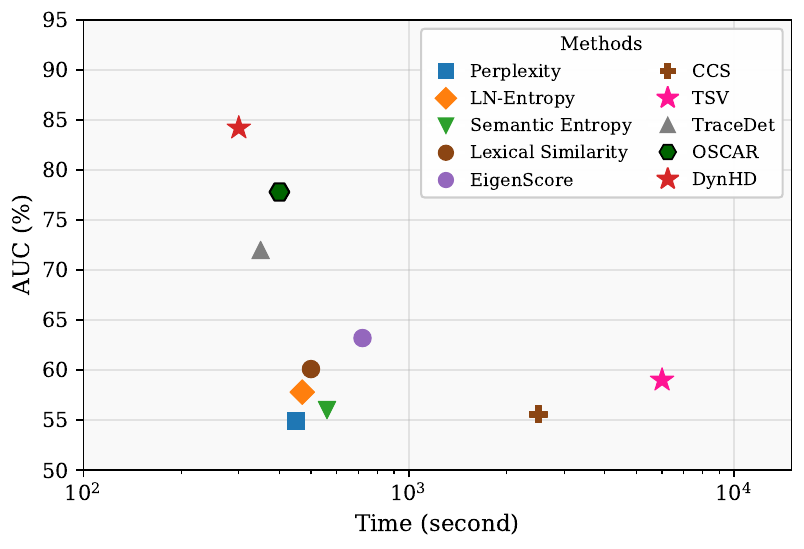}
    \caption{\small\textbf{Pareto frontier.} AUC = LLaDA-8B average. $\star$ = Pareto-optimal. Wall-clock on $4{\times}$H200.}
    \label{fig:pareto_data}
    \vspace{-6em}
\end{wrapfigure}
The key insight: even with $N{=}8$ parallel chains, batched inference keeps the overhead manageable, while the detection + correction synergy produces gains that neither component achieves alone.
\section{Ablation and Analysis}
\label{sec:ablation}
 
We decompose \method's gains to identify which components are load-bearing, how sensitive the method is to its control parameters, and why cross-chain entropy is an effective uncertainty signal.
\subsection{Localization and Correction are Complementary}
\label{sec:ablation_complementarity}
\begin{table}[b]
\centering

\small
\begin{tabular}{@{}lccc@{}}
\toprule
\textbf{Configuration} & \textbf{F1 (\%)} & \textbf{AUROC} & $\Delta$\textbf{F1} \\
\midrule
Unguided decoding & 62.9 & --- & --- \\
Best-of-N selection (no correction) & 62.9 & 82.7 & +0.0 \\
Correction only (random spans) & 62.8 & --- & $-$0.1 \\
Majority vote --- token & 65.1 & --- & +2.2 \\
Localization + untargeted correction & 65.4 & 83.1 & +2.5 \\
\textbf{\method{} (loc.\ + targeted)} & \textbf{69.0} & \textbf{86.5}\textsuperscript{$\dagger$} & \textbf{+6.1} \\
\bottomrule
\multicolumn{4}{@{}l}{\scriptsize \textsuperscript{$\dagger$} Under LLM-as-Judge. EM-eval AUROC = 76.4.}
\end{tabular}
\caption{\small \textbf{Complementarity ablation} (LLaDA-8B, QA macro-avg).
Localization-only AUROC = 82.7 (raw entropy signal, no correction).}
\label{tab:complementary}
\end{table}
\begin{table}[t]
\centering
\small
\begin{tabular}{@{}lcccccc@{}}
\toprule
\textbf{Order} & \textbf{Triv.\ F1} & \textbf{CQ F1} & \textbf{HQA F1}
    & \textbf{RT R-L} & \textbf{AUROC} & \textbf{Avg F1} \\
\midrule
All learned              & 56.8 & 85.4 & 43.1 & 33.8 & 0.541 & 61.8 \\
All random               & 61.2 & 85.4 & 47.3 & 37.4 & 0.608 & 64.6 \\
Hybrid (50/50)           & 64.8 & 85.4 & 49.8 & 40.1 & 0.627 & 66.7 \\
Entropy-ordered          & 59.4 & 85.4 & 44.7 & 35.2 & 0.591 & 63.2 \\
\textbf{\method{} (1L+7R)} & \textbf{68.9} & \textbf{85.4} & \textbf{52.7}
    & \textbf{43.2} & \textbf{0.639} & \textbf{69.0} \\
\bottomrule
\end{tabular}
\caption{\small\textbf{Demasking order ablation} (LLaDA-8B). \method{} default = 1L+7R.}
\label{tab:order}
\end{table}
Table~\ref{tab:complementary} shows that localization alone does not affect F1 performance, while random span corrections slightly reduce results. The complete \method{} pipeline integrates localization with targeted correction, achieving a 6.1-point F1 improvement and 86.5 AUROC, outperforming untargeted correction (+2.5 F1) and token-level majority voting (+2.2 F1). Table~\ref{tab:order} shows the default \method{} configuration (1L+7R) outperforms alternative ordering schemes across all metrics. These results demonstrate the synergy between localization and targeted correction within \method{}, validating its ability to enhance model output quality.
 \subsection{Sensitivity Analysis}
\label{sec:ablation_sensitivity}
Figure~\ref{fig:nchains} shows the effect of chains $N \in \{1, 2, 4, 8, 16, 32\}$ on LLaDA-8B (QA macro-avg), where F1 gain and AUROC increase from $N=1$ to $N=4$, peak at $N=8$, and plateau at $N=16$–$32$; thus, $N=8$ provides optimal localization quality versus cost. For demasking order, the hybrid schedule (one learned step followed by $(T_r - 1)$ random steps) 
\begin{wrapfigure}{r}{0.5\textwidth}
    \centering
    \includegraphics[width=0.48\textwidth]{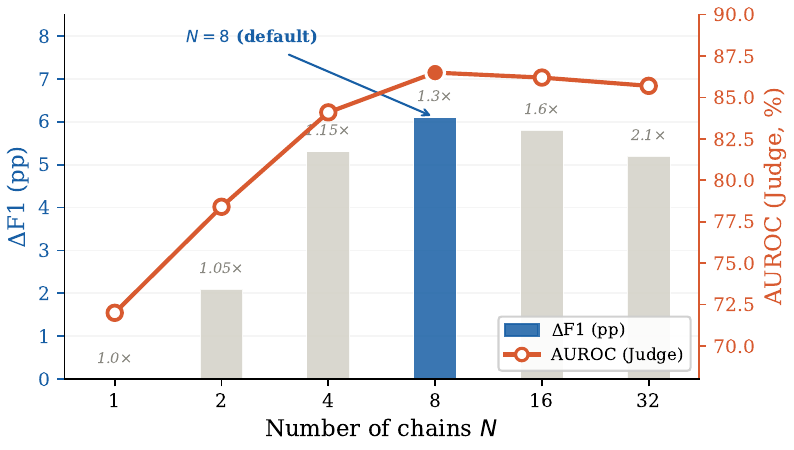}
    \caption{\small Effect of chains $N$ on LLaDA-8B (QA macro-avg).}
    \label{fig:nchains}
    \vspace{-3em}
\end{wrapfigure}
outperforms alternatives: all-learned achieves 61.8 F1 (below 62.9 baseline), all-random reaches 64.6, and entropy-ordered yields 63.2. The learned step anchors the most confident token, while random steps add diversity as in Table~\ref{tab:order}. 
\method corrections show 91–97\% precision in improving outputs (57 vs. 5 on TriviaQA, 34 vs. 1 on HotpotQA; Appendix~\ref{app:precision}). The threshold $\alpha$ shows optimal F1 in $[0.15, 0.25]$ across $\{0.05, 0.10, \ldots, 0.40\}$, degrading at extremes (Appendix~\ref{app:alpha}).
\vspace{-0.em}
\par
\subsection{Why Does Cross-Chain Entropy Work?}
\label{sec:analysis}
Cross-chain entropy leverages the architectural difference between autoregressive (AR) and diffusion language models (DLMs). In AR decoding, uncertainty at position \(i\) is hidden at \(i+1\) since only the argmax token propagates, collapsing the distribution. Recovering uncertainty requires independent resamples, mixing sampling variance with factual uncertainty. DLMs maintain the full probability distribution, sampling various paths via different reveal orders. Cross-chain entropy measures the width of this path distribution at each position, capturing a factual uncertainty signal absent in AR models. This unsupervised signal from \(N=8\) DLM chains outperforms trained classifiers reconstructing commitment uncertainty from trajectory features, confirmed by ablations. Unlike TraceDet and DynHD, which learn from hallucination labels, cross-chain entropy directly measures this signal, surpassing trained classifiers without supervision. Confident-but-wrong cases, where chains agree on incorrect answers (\(H_\times=0\)), represent 9–20\% of hallucinated positions (Appendix~\ref{app:cbw}), indicating a need for retrieval augmentation rather than remasking, as shown by zero-entropy readout.
\section{Conclusion}
We introduced commitment uncertainty localization to identify unreliable token
positions from diffusion language models' denoising trajectories, formalized
through the CDH metric. OSCAR, our training-free framework, runs $N$ parallel
chains with randomized reveal orders, computes cross-chain entropy to localize
unstable positions, and corrects them via targeted remasking. On LLaDA-8B and
Dream-7B, OSCAR improves F1 by $+8.3 \pm 0.4$, reduces hallucinated spans by
41.1\%, and achieves 86.5 AUROC, surpassing trained classifiers at
$1.3{\times}$ overhead. The signal's effectiveness stems from a structural
asymmetry: DLMs maintain full distributions throughout denoising, exposing
commitment-order effects that AR models collapse at each step. Crystallization
analysis reveals task-dependent timing, supporting early-exit detection for
reducing overhead. Our findings demonstrate that diffusion language models
contain native uncertainty signals enabling both error localization and
correction during inference---an affordance absent in autoregressive generation.

\paragraph{Future work.}
Several directions follow naturally from this work.
First, the crystallization analysis suggests that much of the useful uncertainty
signal emerges in the earliest denoising steps; this raises the possibility of
early-exit detection and lower-cost correction schedules.
Second, extending commitment uncertainty localization beyond discrete masked
diffusion to continuous or hybrid diffusion language models would test how
general the signal is across architectures.
Third, adaptive intervention policies---including task-dependent thresholds
$\alpha$, span-selection strategies, or correction schedules---may improve the
accuracy--efficiency trade-off further.
Finally, integrating OSCAR-style localization with preference optimization,
retrieval planning, or external tool use could broaden its role from factual
repair to more general inference-time control.
We release the trajectory divergence toolkit to support future work on
localization, correction, and uncertainty-aware generation in DLMs.

\section*{Limitations}
\label{sec:limitations}
Five limitations bound the current work.
(1)~OSCAR requires running $N$ parallel chains, increasing peak VRAM by
${\sim}1.67{\times}$ for $N{=}8$ (32.1~GB vs.\ 19.2~GB) even though
wall-clock time only increases $1.3{\times}$ via batching.
(2)~For queries where the model lacks relevant knowledge entirely, correction
via remasking cannot help---the model will hallucinate consistently across all
chains.
(3)~The LLM-as-Judge evaluation introduces its own biases.
(4)~We evaluate two DLMs; generalization to future architectures is not
guaranteed.
(5)~Span aggregation hyperparameters ($\pm2$ token extension, 3-token minimum)
are validated empirically but not theoretically grounded.

\section*{Ethics Statement}
This work proposes training-free inference-time techniques for reducing
hallucination in diffusion language models. We do not introduce new training
data, fine-tuned models, or large-scale data collection. The models evaluated
(LLaDA-8B, Dream-7B) are publicly available research artifacts; evaluations
use public benchmarks (TriviaQA, HotpotQA, CommonsenseQA, RAGTruth).
LLaDA-8B and Dream-7B are the only publicly available DLMs at scale as of
early 2026; our evaluation therefore covers the full available ecosystem.
Technical limitations---including computational overhead, failure modes under
complete knowledge absence, evaluation biases, and hyperparameter
sensitivity---are detailed in \S\ref{sec:limitations}. The LLM-as-Judge
evaluation (GPT-4o) introduces potential biases; no human evaluation of
user-perceived helpfulness is included. GPT-4o assisted in polishing this
manuscript; associated costs and biases are acknowledged.

\bibliography{references,oscar_references_complete}
\bibliographystyle{colm2026_conference}


\newpage
\section*{Appendix}
\noindent{\Large\bfseries Table of Contents}\par\vspace{-0.6em}\noindent\rule{\linewidth}{0.4pt}

\noindent\textbf{A\quad RAGTruth: Span-Level Results} \hfill \pageref{app:ragtruth}\vspace{0.2em}
\begin{addmargin}[2em]{0em}
    A.1\enspace Retrieval Augmentation              \dotfill \pageref{app:retrieval}\\[2pt]
    A.2\enspace CDH Localization Curves             \dotfill \pageref{app:cdh}
\end{addmargin}\vspace{0.5em}

\noindent\textbf{B\quad Ablation Studies} \hfill \pageref{app:ablation}\vspace{0.2em}
\begin{addmargin}[2em]{0em}
    B.1\enspace Correction Precision    \dotfill \pageref{app:precision}\\[2pt]
    B.2\enspace Sensitivity: $\alpha$, $T_r$, Span, $N$-Paths          \dotfill \pageref{app:span}\\[2pt]
    B.3\enspace Confident-but-Wrong Analysis          \dotfill \pageref{app:cbw}\\[2pt]
    B.4\enspace Extended Generation Metrics          \dotfill \pageref{app:generation}
    
\end{addmargin}\vspace{0.5em}

\noindent\textbf{C\quad Baseline Comparisons} \hfill \pageref{app:baselines}\vspace{0.2em}
\begin{addmargin}[2em]{0em}
    C.1\enspace Per-Dataset Breakdown       \dotfill \pageref{app:breakdown}\\[2pt]
    C.2\enspace SelfCheckGPT-DLM Comparison                     \dotfill \pageref{app:selfcheck} 
\end{addmargin}\vspace{0.5em}

\noindent\textbf{D\quad Statistical Validation} \hfill \pageref{app:stats}\vspace{0.2em}
\vspace{0.5em}

\noindent\textbf{E\quad Implementation Details} \hfill \pageref{app:impl}\vspace{0.2em}
\begin{addmargin}[2em]{0em}
    E.1\enspace LLM-as-Judge Prompt                              \dotfill \pageref{app:judge}
\end{addmargin}

\vspace{0.3em}\noindent\rule{\linewidth}{0.4pt}

\newpage
\appendix


\section{RAGTruth: Span-Level Results}
\label{app:ragtruth}

Table~\ref{tab:span} provides the full per-subset breakdown behind
Figure~\ref{fig:ragtruth}.
Span reduction is highest on Data2Text (43.7\%), where structured facts
produce clear-cut uncertain positions; the highest fraction of improved
examples is on QA (75.0\%), where single-span corrections have the largest
relative impact.
Positive $\Delta$FS across all subsets confirms that corrections improve
factual precision rather than simply deleting content.

\begin{table}[h]
\centering
\small
\begin{tabular}{@{}lrr ccc@{}}
\toprule
Subset & $N$ & Hall.\% & AUROC & $\Delta$FS & Span Red.\% \\
\midrule
QA        & 788  & 18.4 & 0.612 & +0.052 & 38.4 \\
Summary   & 500  & 24.2 & 0.641 & +0.073 & 41.2 \\
Data2Text & 500  & 66.8 & 0.663 & +0.092 & 43.7 \\
\midrule
All       & 1788 & 33.1 & 0.639 & +0.072 & 41.1 \\
\bottomrule
\end{tabular}
\caption{\small Span-level correction on RAGTruth (LLaDA-8B).
$\Delta$FS: FactScore change (positive = more precise).
Span Red.: reduction in hallucinated span mass.}
\label{tab:span}
\end{table}

\subsection{Retrieval Augmentation}
\label{app:retrieval}

Table~\ref{tab:retrieval} isolates the effect of retrieval strategy.
\method~ without retrieval (60.8) already exceeds naive RAG (53.5) by 7.3
points, the primary gain is from targeted remasking, not retrieved evidence. Adding targeted span-level retrieval yields a further +3.4 over \method~ alone, because evidence is fetched for each uncertain claim rather than the full query.

\begin{table}[h]
\centering
\small
\begin{tabular}{@{}lcc@{}}
\toprule
Configuration & F1 (\%) & $\Delta$ vs.\ Unguided \\
\midrule
Unguided decoding      & 51.7 & ---   \\
Naive RAG              & 53.5 & +1.8  \\
OSCAR (no retrieval)   & 60.8 & +9.1  \\
OSCAR + Naive RAG      & 62.4 & +10.7 \\
OSCAR + Targeted RAG   & 64.2 & +12.5 \\
\bottomrule
\end{tabular}
\caption{Retrieval comparison (LLaDA-8B, TriviaQA + HotpotQA avg F1).}
\label{tab:retrieval}
\end{table}

\subsection{CDH Localization Curves}
\label{app:cdh}

\begin{wrapfigure}{r}{0.5\linewidth} 
    \centering
    \includegraphics[width=\linewidth]{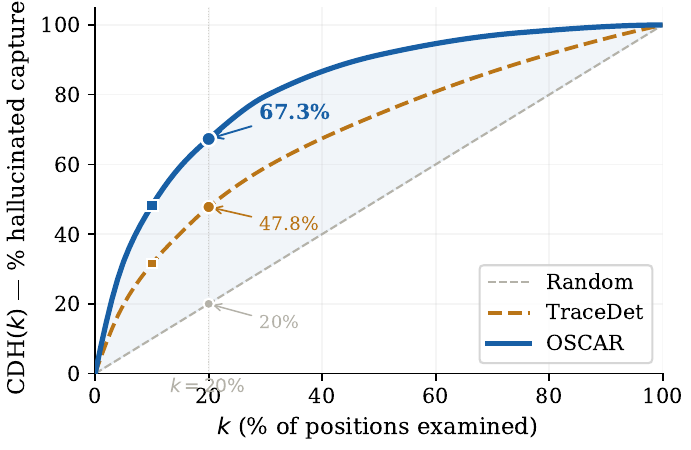}
    \caption{\small CDH$(k)$ on RAGTruth. At $k{=}20\%$, OSCAR captures 67.3\% of hallucinated positions vs.\ 47.8\% (TraceDet) and 20\% (random).}
    \label{fig:cdh}
\end{wrapfigure}

Figure~\ref{fig:cdh} shows that hallucinated positions concentrate in the high-entropy tail: examining only the top 20\% of positions by cross-chain entropy captures two-thirds of all true errors, making targeted remasking practical without modifying the remaining 80\% of tokens. At $k{=}10\%$, OSCAR already captures 48.2\% (vs.\ 31.5\% TraceDet, 10\% random), confirming that the unsupervised signal localizes hallucinations more effectively than trained alternatives.

\section{Ablation Studies}
\label{app:ablation}

\subsection{Correction Precision}
\label{app:precision}

Among examples where OSCAR applies a correction, 91.9\% are improved on
TriviaQA (57 corrected, 5 broken out of 500) and 97.1\% on HotpotQA
(34 corrected, 1 broken).
CommonsenseQA triggers no intervention.

\subsection{Sensitivity: $\alpha$, $T_r$, Span, $N$-Paths}
\label{app:alpha}
\label{app:tr}
\label{app:span}
\label{app:npaths}

\begin{table}[h]
\label{tab:alpha}
\begin{tabular}{@{}lccc@{}}
\toprule
$\alpha$ & \% Remasked & F1 (\%) & AUROC \\
\midrule
0.05 & 5\%  & 64.1 & 83.8 \\
0.10 & 10\% & 66.5 & 85.1 \\
0.15 & 15\% & 68.2 & 86.0 \\
\textbf{0.20} & \textbf{19\%} & \textbf{69.0} & \textbf{86.5} \\
0.25 & 25\% & 68.1 & 86.3 \\
0.30 & 30\% & 66.8 & 85.9 \\
0.40 & 40\% & 63.7 & 84.2 \\
\bottomrule
\end{tabular}
\centering\small
\caption{\small Threshold $\alpha$ sensitivity (LLaDA-8B, QA macro-avg).}
\end{table}

\begin{table}[h]
\centering\small
\begin{tabular}{@{}p{3.8cm} cc@{}}
\toprule
Setting & F1 (\%) / $\Delta$FS & Cost \\
\midrule
\multicolumn{3}{@{}l}{\textit{Refinement steps $T_r$}} \\
$T_r{=}2$                        & 65.4 / +0.041 & 0.03$\times$ \\
$T_r{=}4$                        & 67.8 / +0.058 & 0.06$\times$ \\
$\mathbf{T_r{=}8}$               & \textbf{69.0 / +0.072} & \textbf{0.10$\times$} \\
$T_r{=}16$                       & 69.2 / +0.074 & 0.19$\times$ \\
\midrule
\multicolumn{3}{@{}l}{\textit{Span aggregation ($w$, $\ell_{\min}$)}} \\
$w{=}0$, $\ell{=}1$              & 65.8 & 28.4\% red. \\
$w{=}1$, $\ell{=}3$              & 67.4 & 35.2\% \\
$\mathbf{w{=}2, \ell{=}3}$       & \textbf{69.0} & \textbf{41.1\%} \\
$w{=}3$, $\ell{=}3$              & 68.3 & 45.8\% \\
$w{=}2$, $\ell{=}5$              & 67.9 & 37.6\% \\
\midrule
\multicolumn{3}{@{}l}{\textit{$N$-paths FactScore (RAGTruth)}} \\
$N{=}1$ & $\Delta$FS $= -0.030$ & --- \\
$N{=}4$ & $+0.073$ & --- \\
$\mathbf{N{=}8}$ & $\mathbf{+0.148}$ & --- \\
$N{=}16$ & $+0.048$ & --- \\
\bottomrule
\end{tabular}
\caption{\small Refinement steps $T_r$, span aggregation, and $N$-paths FactScore
(LLaDA-8B). Defaults bolded.}
\label{tab:combined_sensitivity}
\end{table}

$\alpha$ plateaus across $[0.15, 0.25]$; F1 and $\Delta$FS plateau at
$T_r{=}8$; span defaults ($w{=}2$, $\ell_{\min}{=}3$) achieve peak F1;
FactScore peaks at $N{=}8$.

\subsection{Confident-but-Wrong Analysis}
\label{app:cbw}
 
\begin{table}[h]
\centering\small
\begin{tabular}{@{}lrrrc@{}}
\toprule
Dataset & Total Hall. & CBW & Detectable & CBW Rate \\
\midrule
TriviaQA & 245 & 38 & 207 & 15.5\% \\
HotpotQA & 315 & 63 & 252 & 20.0\% \\
RAGTruth & 592 & 53 & 539 &  9.0\% \\
\bottomrule
\end{tabular}
\caption{\small CBW hallucinations (LLaDA-8B). CBW = $H_{\times,i}{=}0$ but
factually wrong.}
\label{tab:cbw}
\end{table}
 
80--91\% of hallucinated positions expose non-zero cross-chain entropy and are
detectable; the remaining CBW cases represent knowledge gaps requiring
retrieval rather than remasking.
\subsection{Extended Generation Metrics}
\label{app:generation}

\begin{table}
\centering\small
\begin{tabular}{@{}l cc cc cc cc@{}}
\toprule
& \multicolumn{2}{c}{EM} & \multicolumn{2}{c}{F1} 
& \multicolumn{2}{c}{R-L} & \multicolumn{2}{c}{BLEU} \\
\cmidrule(lr){2-3}\cmidrule(lr){4-5}\cmidrule(lr){6-7}\cmidrule(lr){8-9}
Dataset & Bef. & Aft. & Bef. & Aft. & Bef. & Aft. & Bef. & Aft. \\
\midrule
\multicolumn{9}{@{}l}{\textit{LLaDA-8B}} \\
TriviaQA & 51.0 & 61.4 & 58.2 & 68.9$\pm$0.3 & 55.1 & 65.4 & 33.2 & 41.8 \\
HotpotQA & 37.0 & 43.6 & 45.2 & 52.7$\pm$0.4 & 42.8 & 50.1 & 24.6 & 30.1 \\
CommQA   & 85.4 & 85.4 & 85.4 & 85.4$\pm$0.0 & 85.4 & 85.4 & 72.1 & 72.1 \\
\midrule
\multicolumn{9}{@{}l}{\textit{Dream-7B}} \\
TriviaQA & 42.0 & 53.8 & 50.5 & 61.2$\pm$0.5 & 48.3 & 58.7 & --- & --- \\
HotpotQA & 30.2 & 37.4 & 38.8 & 46.1$\pm$0.6 & 36.5 & 43.7 & --- & --- \\
CommQA   & 81.2 & 81.2 & 81.2 & 81.2$\pm$0.0 & 81.2 & 81.2 & --- & --- \\
\bottomrule
\end{tabular}
\caption{\small Full generation metrics before/after OSCAR correction
(mean $\pm$ std over 3 seeds).}
\label{tab:generation_full}
\end{table}
\section{Baseline Comparisons}
\label{app:baselines}

\subsection{Per-Dataset Breakdown}
\label{app:breakdown}

\begin{table}[htbp]
\centering\small
\begin{tabular}{@{}lccc@{}}
\toprule
Method & TrivQA & HotQA & Category \\
\midrule
Perplexity       & 52.1 & 50.8 & Output \\
LN-Entropy       & 56.3 & 53.4 & Output \\
Semantic Entropy  & 54.7 & 52.1 & Output \\
Lexical Sim.     & 61.8 & 58.7 & Output \\
EigenScore       & 65.4 & 61.2 & Latent \\
TSV              & 58.2 & 56.8 & Latent \\
\midrule
OSCAR (EM)       & 80.3 & 71.5 & Cross-chain \\
OSCAR (Judge)    & 89.7 & 86.7 & Cross-chain \\
\bottomrule
\end{tabular}
\caption{Fair-baseline AUROC (\%) under unified protocol (LLaDA-8B).}
\label{tab:fair}
\end{table}

All baselines re-evaluated on our DLM outputs under the same pipeline.

\subsection{SelfCheckGPT-DLM Comparison}
\label{app:selfcheck}

\begin{table}[htbp]
\centering\small
\begin{tabular}{@{}lccc@{}}
\toprule
Method & TrivQA & HotQA & Avg \\
\midrule
SelfCheck (overlap)  & 59.3 & 56.1 & 57.7 \\
SelfCheck (entropy)  & 54.8 & 52.4 & 53.6 \\
\midrule
OSCAR (EM)           & 80.3 & 71.5 & 76.4 \\
OSCAR (Judge)        & 89.7 & 86.7 & 86.5 \\
\bottomrule
\end{tabular}
\caption{\small SelfCheckGPT-DLM vs.\ OSCAR (LLaDA-8B, $N{=}8$).}
\label{tab:selfcheck}
\end{table}

OSCAR outperforms the best SelfCheckGPT variant by 18.7 AUROC points;
reveal-order diversification exposes a qualitatively different signal than
independent resampling.

\section{Statistical Validation}
\label{app:stats}

\begin{table}[htbp]
\centering\small
\begin{tabular}{@{}lrcc@{}}
\toprule
Dataset & AUROC & 95\% CI Low & High \\
\midrule
TriviaQA (Judge) & 89.7 & 86.8 & 92.3 \\
HotpotQA (Judge) & 86.7 & 83.2 & 89.9 \\
CommQA (Judge)   & 83.4 & 80.1 & 86.5 \\
Average (EM)     & 76.4 & 73.8 & 79.1 \\
\bottomrule
\end{tabular}
\caption{\small Bootstrap 95\% CIs (1{,}000 resamples, LLaDA-8B).}
\label{tab:bootstrap}
\end{table}

All Judge CIs on TriviaQA and HotpotQA exclude DynHD's LLM-as-judge score (84.2).

\newpage
\section{Implementation Details}
\label{app:impl}

\paragraph{LLM-as-Judge prompt.}
\label{app:judge}
Applied uniformly to all methods via GPT-4o:

\begin{quote}\small\ttfamily
You are evaluating whether an AI-generated answer is factually correct.
The answer does not need to match the reference exactly---it should be
semantically equivalent and contain the key facts.
Rate as CORRECT if the core factual content matches, INCORRECT if it
contains factual errors, PARTIAL if partially correct.
Respond with only the label.
\end{quote}
\end{document}